# Character Distributions of Classical Chinese Literary Texts: Zipf's Law, Genres, and Epochs

[†]Chao-Lin Liu   [‡]Shuhua Zhang   [*]Yuanli Geng   [⁑]Huei-ling Lai   [↑]Hongsu Wang

[†‡*↑]Harvard University, USA
[†⁑]National Chengchi University, Taiwan
[‡]Northwestern Polytechnical University, China
[*]Liaoning University, China

{[†]chaolin, [⁑]hllai}@nccu.edu.tw, [‡]shuhuazhang1@sina.com, [*]geng99999@126.com, [↑]hongsuwang@fas.harvard.edu

## Introduction and Main Findings

Chinese characters are the basic units for Chinese words, and a Chinese word can include one, two, or more characters. Many characters can function as words in Chinese. For instance, "人文" [1] represents "humanities", and "人" and "文" are characters that carry their own meanings. Words that contain two, three, and more characters can be referred as *bigrams*, *trigrams* and other appropriate names for *n-grams* [2]. Chinese does not separate words with spaces like English, so a reader must "segment" a character string into words to understand Chinese statements.

Mandarin Chinese has evolved over the past thousands of years [3]. Documents written in current vernacular Chinese contain a large number of bigrams and trigrams, while texts of classical Chinese [4] contain many more unigrams. In the ASBC [5], unigrams contribute only 2.1% of word types, but constitute a 44.8% share of word tokens [6]. In contrast, bigrams and trigrams together constitute 82% and 53% of the word types and tokens, respectively.

Zipf discovered that the relative frequencies of words are inversely proportional to their ranks in Chinese and English documents [7] in his endeavor to establish a theory of *Principle of Least Effort* [8]. Many researchers have attempted to find the parameters in the Zipf-Mandelbrot distributions [9] for English and for Chinese [10], and many have explored extensions and applications of the law [11].

Previous research works for Chinese have focused mainly on fitting the Zipfian distributions to Chinese corpora. Some considered a distributed sample of corpora [12], and others confined their analysis to a specific corpus [11]. Some researchers have pondered on explanations for why the statistics of languages obey the Zipfian distributions [13,14].

We examined the Zipfian distributions of 14 collections of Chinese texts that were published from 1046BC [15] to 2007AD, and we found that the genres and epochs of the collections influence the distributions. The majority of our collections are poetic works written in classical Chinese. We also included official documents of the Tang Dynasty, novels of the Ming and Qing dynasties, and news articles of modern days.

The character distributions for the corpora of poems of 618-1644AD [16] exhibit strikingly similar Zipfian distributions. In contrast, the character distributions of the three genres of corpora that were all written in the Tang dynasty are distinguishable, although the characters frequently which are used in these documents are very similar. The word distribution of the ASBC differs significantly from other character distributions, indicating the importance of differentiating character- and word-based models of Chinese.

**Table 1. The corpora used in this study include texts published during 1046BC-2007AD.**

| Acronym | Collection | Time | Acronym | Collection | Time |
|---|---|---|---|---|---|
| **SJ** | 詩經 | 1046-476BC | **CV** | 楚辭 | 475-221BC |
| **HF** | 漢賦(文選) | 202BC-420AD | **PT** | 先秦漢魏晉南北朝詩 | Before 589AD |
| **CTW** | 全唐文 | 618-907AD | **MZM** | 唐墓誌銘 | 618-907AD |
| **CTP** | 全唐詩 | 618-907AD | **CSP** | 全宋詩 | 960-1279AD |
| **CSL** | 全宋詞 | 960-1279AD | **YSX** | 元詩選 | 1271-1368AD |
| **LCSJ** | 列朝詩集 | 1368-1644AD | **JTTW** | 西遊記 | ca. 16$^{th}$ century |
| **DRC** | 紅樓夢 | ca. 18$^{th}$ century | **ASBC** | 平衡語料庫 | 1981-2007AD |

## Corpora and Comparisons

For ease of references, we assigned an acronym for each of the 14 corpora, and show their names in Chinese (**Collection**) and periods of publication (**Time**) in Table 1 [17].

The corpora consist of representative literature that has been published since 1046BC. In particular, we have at least one collection for each of the major dynasties that existed before 1644AD. The majority of our collections are of poetic works, which fact lends itself to the study of the effects of genres on the character distributions [18]. A collection of poetic works for the Qing Dynasty [19] is unavailable because an editorial committee is still working on its production [20].

The corpora contain more than 42 million characters, excluding the punctuation marks that were added into the corpora by the data providers. When counting the characters, we also exclude characters that cannot be shown on ordinary computers. The frequencies of such rare and obsolete characters are not large, so ignoring them will not affect the statistical properties reported in this study.

Only the ASBC was segmented and the segmentation was verified by human experts. Hence, we can inspect its character and word distributions. The other corpora were written in classical Chinese and we do not have a reliable way for segmentation, so we will only analyze the character distributions.

We created charts that are based on the typical form of Zipf's law:

$$\log\left(\frac{f(w)}{N}\right) = k - \alpha \log(r(w)), \tag{1}$$

where $w$, $f(w)$, and $r(w)$ denote a word, its frequency, and rank in a corpus, respectively. The rank of the most frequent word in a corpus is 1. $N$ is the size of the corpus, and $k$ and $\alpha$ are constants.

**Table 2. The most frequent ten characters in poems and lyrics remaining stable over time**

| Corpus | Ranks | | | | | | | | | |
|---|---|---|---|---|---|---|---|---|---|---|
| | 1 | 2 | 3 | 4 | 5 | 6 | 7 | 8 | 9 | 10 |
| PT | 不 | 無 | 風 | 有 | 人 | 雲 | 之 | 何 | 日 | 我 |
| CTP | 不 | 人 | 山 | 無 | 風 | 一 | 日 | 雲 | 有 | 何 |
| CSP | 不 | 人 | 一 | 無 | 山 | 有 | 風 | 來 | 天 | 日 |
| CSL | 人 | 風 | 花 | 一 | 不 | 春 | 無 | 雲 | 來 | 天 |
| YSX | 不 | 人 | 山 | 風 | 一 | 雲 | 天 | 日 | 有 | 無 |
| LCSJ | 不 | 人 | 風 | 山 | 一 | 花 | 日 | 雲 | 有 | 無 |

## Observations and Discussions: Influences of Genres and Epochs

The generalizability of the Zipf's law is the main reason that it has attracted the attention of many researchers. It can be applied to various natural distributions including those of part-of-speech of words [21], city sizes [22], and corporal revenues [23].

Figure 1 shows the Zipfian curves when we consider the character distributions of all of the 14 corpora. We intentionally plot the curves in one chart, although this makes the individual curves undistinguishable. Although the curves are not linear, which is common as reported in the literature, the curves show a consistent trend, suggesting a common regularity that is shared by Chinese texts that were produced over the period of 3000 years.

Instead of treating the 14 corpora as a single corpus to fit the resulting distribution for Zipf's law, we examined the curves to investigate possible factors that influenced the positions of the curves. In Figure 2, we show the curves of lyrics ("詞" /ci2/) and poems ("詩" /shi1/. The left halves of the curves overlap almost perfectly, which strongly indicates that the poetic works share very close statistical characteristics.

Table 2 lists ten most frequent characters found in each of the corpora and for which the curves are plotted in Figure 2. The lists are very similar, and, out of the 60 characters in Table 2, there are only 16 distinct characters [24]. In fact, we can compare the most frequent characters of any two corpora, e.g., the CTP and the CSP, to further investigate their similarity [25], and we found that the most frequent 1700 characters in the CTP and the CSP are the same characters.

Not all of the corpora of poetic works have similar curves. We added the curves for the SJ, CV, and HF in Figure 3, and it is evident that these new curves do not overlap with those in Figure 2 very well. The poetic works in the SJ, CV, and HF were produced very much earlier than those listed in Figure 2.

While the time of the production of the corpora affects the Zipfian curves, the curves for corpora that were produced in the same dynasty may not be the same. The CTW, MZM, and CTP are three different types of works that were all produced in the Tang Dynasty. We compared the most frequent characters shared by the CTP and CTW, and found that the sets of their most frequent 2000 characters differ only in three characters. Despite such an extreme overlap, their curves in Figure 4 suggest that genre affects the character distributions.

Given the above observations, one may have expected that the curves for the novels that were published in the 16[th] and 18[th] centuries, i.e., the JTTW and DRC, will deviate from the curves for the earlier poems, as the curves in Figure 5 show.

## Character vs. Word distributions

We considered the character distribution when we analyzed the contents of the ASBC in Figure 1, where we found that the character distributions of the vernacular and classical Chinese texts show a reasonable common trend. The ASBC contains documents that were written in vernacular Chinese, so we must also analyze its word distribution, and Figure 6 shows the curves for both the character and word distributions.

A chart like that of Figure 6 can mislead one to infer that Chinese texts do not conform to Zipf's law. It is well accepted that the number of Chinese characters is limited, although there is no consensus about the exact number of characters. In contrast, there is virtually no limit on the number of legal Chinese n-grams. As a result, the sharp downturn of the character distribution and the intersection of the two curves in Figure 6 are expected, and this can be observed in languages other than Chinese in some special settings [26]. We should examine the Zipfian curves on the same basis, e.g., character or word, while considering cultural factors that may influence actual language usage.

## Concluding Remarks

We have judged the similarity between the Ziphian curves based on the visual closeness, though we can quantify the degree of similarity when desired [27]. Researchers have noticed the deviations of Zipfian curves at the high- and low-frequency ends [27,28], and tried to find density functions that fit the data. The statistics at the high-frequency ends of the curves are evidently more reliable. We focused on the deviations at the high-frequency ends of the curves, and discussed how the deviations in these regions may relate to the genres and epochs of the corpora, employing the lists of most frequent characters of the corpora as extra supports.

## Acknowledgments

This research was supported in part by the grant 104-2221-E-004-005-MY3 from the Ministry of Science and Technology of Taiwan, the grant USA-HAR-105-V02 from the Top University Strategic Alliance of Taiwan, and the Senior Fulbright Research Grant 2016-2017.

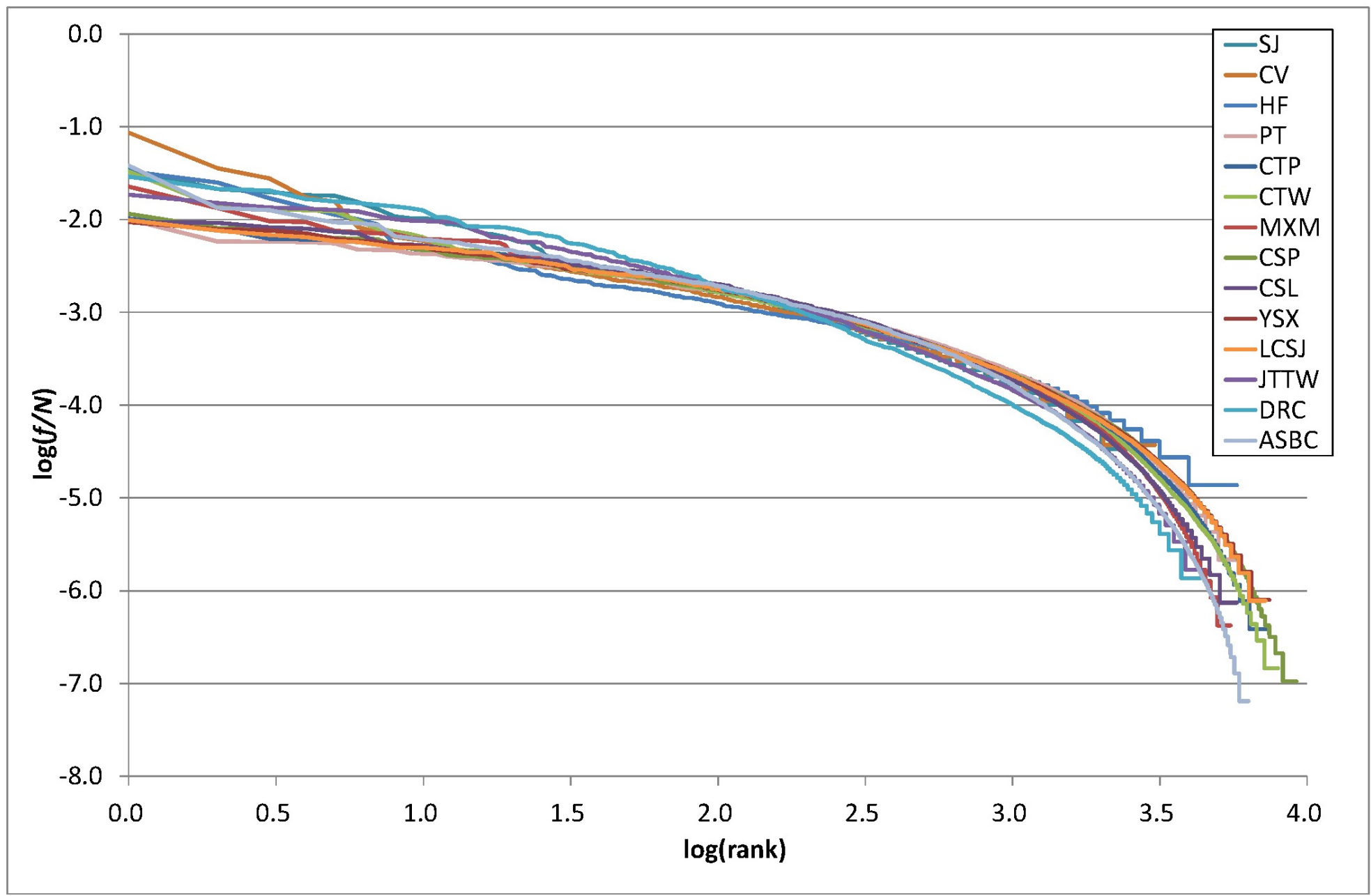

**Figure 1. Zipfian curves of 14 corpora suggest a common trend. (Character distributions)**

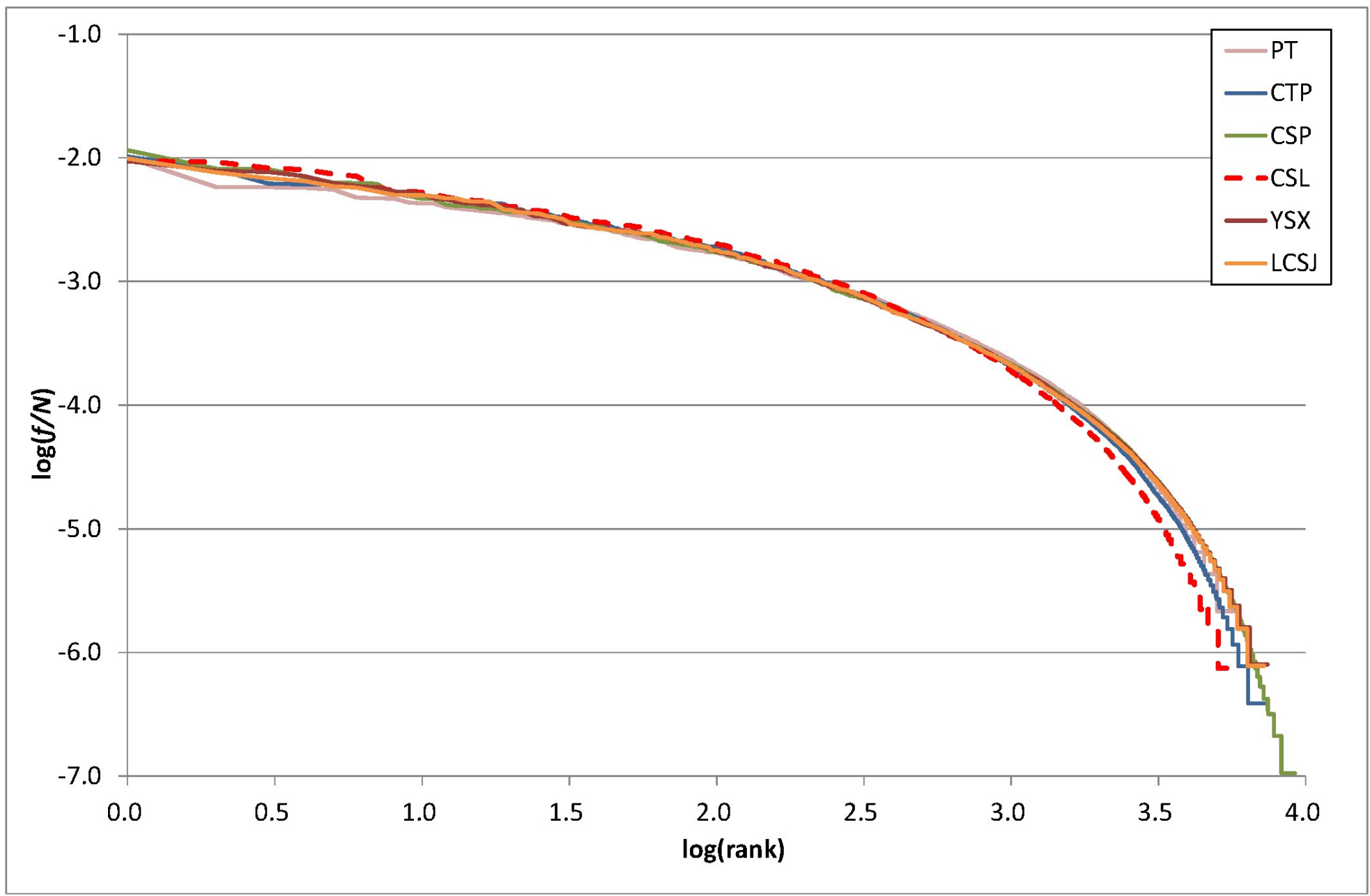

Figure 2. Zipfian curves of the corpora of lyrics (the red dashed curve) and poems (the rest) are strikingly similar. (Character distributions)

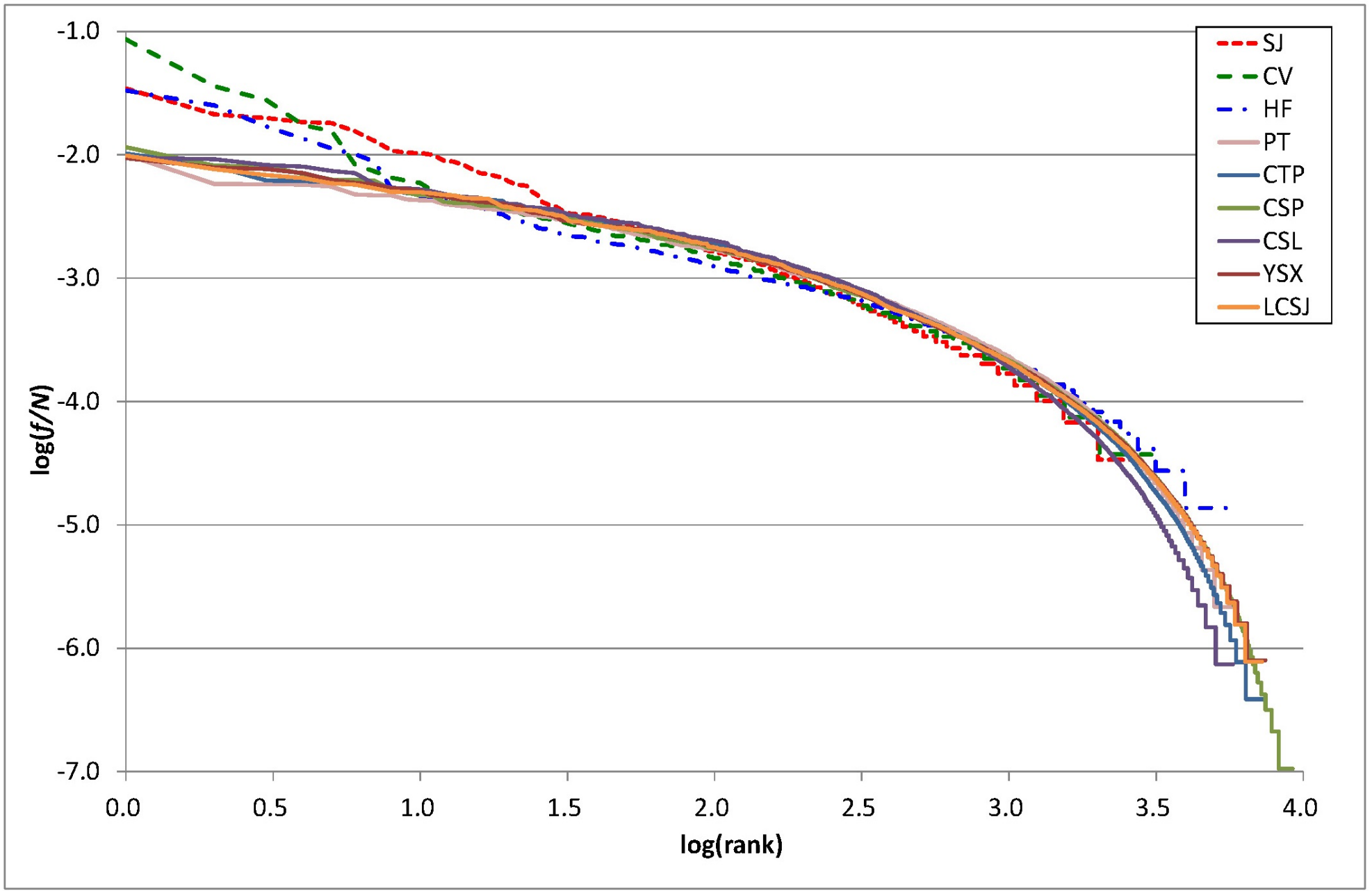

**Figure 3.** Curves for ancient poetic works (SJ, CV, and HF) do not coincide with those of later poems.

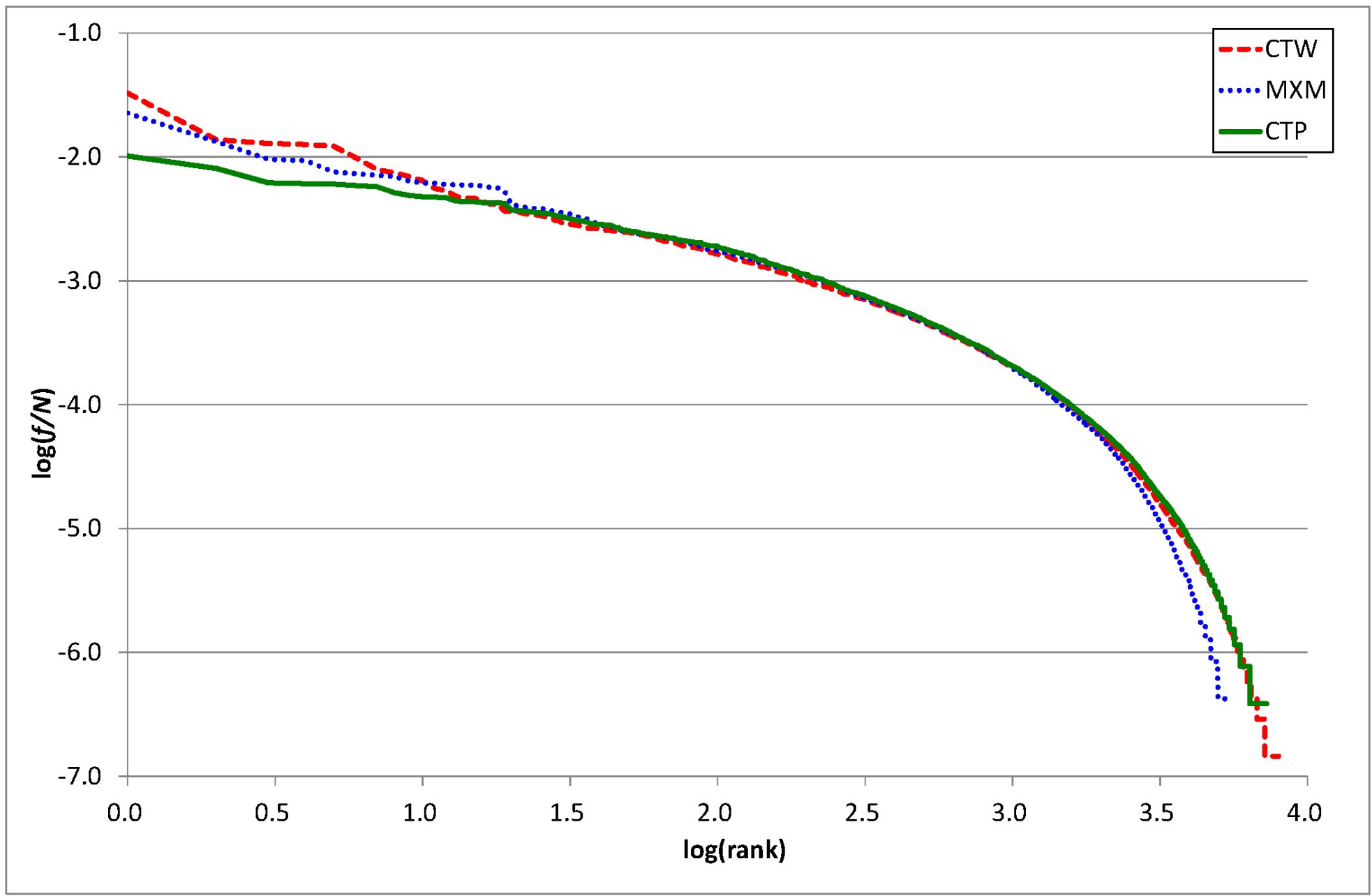

Figure 4. Curves of corpora that belong to the same dynasty but of different genres deviate from each other.

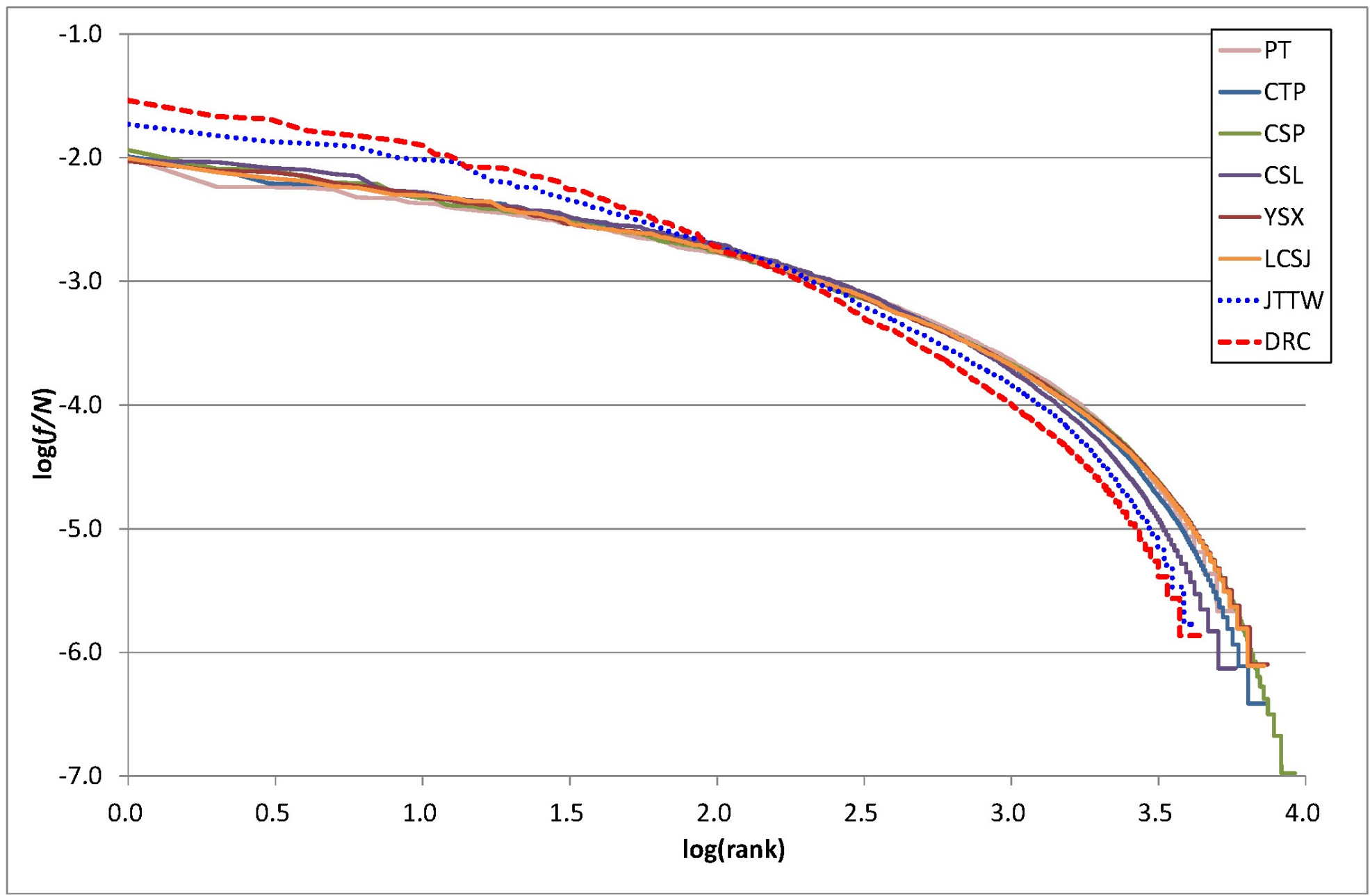

**Figure 5. Curves of corpora that contain novels of 16[th] and 18[th] century (JTTW & DRC, respectively) deviate from the curves in Figure 2.**

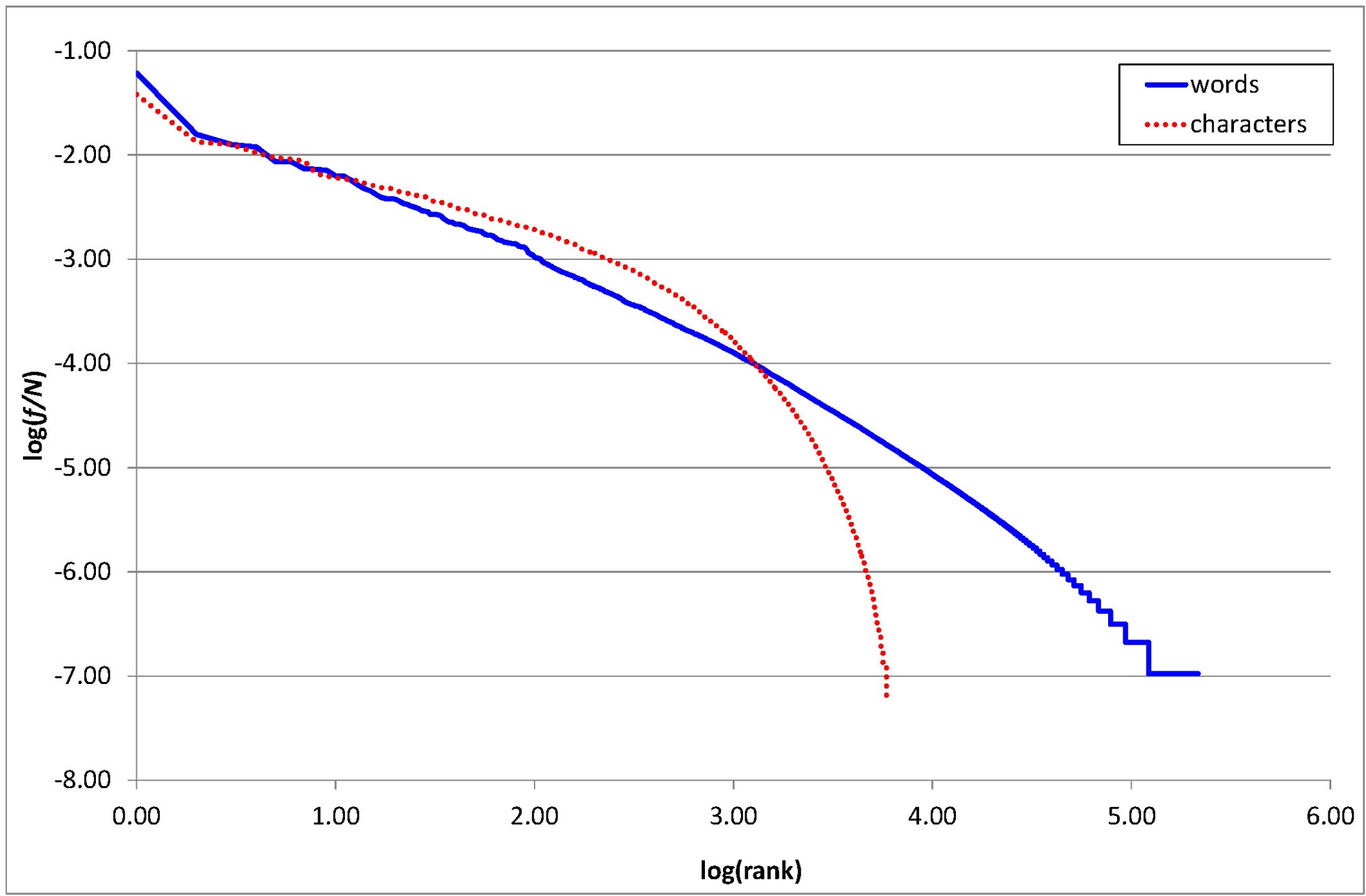

**Figure 6. Word and character distributions for the ASBC differ significantly.**